\documentclass[journal,twoside,web]{ieeecolor}
\usepackage{generic}
\usepackage{cite}
\usepackage{subfig}
\usepackage{amsmath,amssymb,amsfonts}
\usepackage{graphicx}
\usepackage[ruled,linesnumbered]{algorithm2e}
\usepackage{hyperref}
\hypersetup{hidelinks}
\usepackage{textcomp}
\usepackage{float}
\usepackage{comment}
\usepackage{multirow}
\usepackage{multicol}
\usepackage{booktabs}
\usepackage{verbatim}
\usepackage{mathtools}
\usepackage{algpseudocode}
\usepackage{pifont}
\let\oldnl\nl 
\newcommand{\nonl}{\renewcommand{\nl}{\let\nl\oldnl}}
\def\BibTeX{{\rm B\kern-.05em{\sc i\kern-.025em b}\kern-.08em
    T\kern-.1667em\lower.7ex\hbox{E}\kern-.125emX}}
\begin{document}
\title{Long-Term Alzheimer’s Disease Prediction: A
Novel Image Generation Method Using Temporal
Parameter Estimation with Normal Inverse
Gamma Distribution on Uneven Time Series}

\author{Xin Hong, Xinze Sun, Yinhao Li, and Yen-Wei Chen
\thanks{This work is supported in part by the Natural Science Foundation of Fujian Province, China under Grant 2022J01318; in part by the Scientific Research Start-up Fund Project for High-level Researchers of Huaqiao University under Grant 22BS105. }
\thanks{Xin Hong is with the College of Computer Science and Technology, Huaqiao University, Xiamen 361021, China. 
    and also with the Key Laboratory of Computer Vision and Machine Learning in Fujian Province, Xiamen 361021, China. 
    (Corresponding author,email: xinhong@hqu.edu.cn; 10409035@qq.com). }
\thanks{Xinze Sun is with the College of Computer Science and Technology, Huaqiao University, Xiamen 361021, China. 
    (email: 1095948381@qq.com).}
\thanks{Yinhao Li is with the college of Information Science and Engineering,Ritsumeikan University, Ibaraki 567-8570, Japan. 
    (email: yin-li@fc.ritsumei.ac.jp).}
\thanks{Yen-Wei Chen is with the college of Information Science and Engineering,Ritsumeikan University, Ibaraki 567-8570, Japan. 
    (email: chen@is.ritsumei.ac.jp).}}

\maketitle

\begin{abstract}
Image generation can provide physicians with a foundation for imaging diagnosis in the prediction of Alzheimer's Disease (AD).
Recent research has shown that long-term AD predictions using image generation often face difficulties in maintaining disease-related characteristics when dealing with irregular time intervals in sequential data. 
Considering that the time-related aspects of the distribution can reflect changes in disease-related characteristics when images are distributed unevenly, this research proposes a model to estimate the temporal parameter within the Normal Inverse Gamma Distribution (T-NIG) to assist in generating images over the long term. 
The T-NIG model employs brain images from two different time points to create intermediate brain images, forecast future images, and predict the disease.  T-NIG is designed by identifying features using coordinate neighborhoods. It incorporates a time parameter into the normal inverse gamma distribution to understand how features change in brain imaging sequences that have varying time intervals. Additionally, T-NIG utilizes uncertainty estimation to reduce both epistemic and aleatoric uncertainties in the model, which arise from insufficient temporal data.
In particular, the T-NIG model demonstrates state-of-the-art performance in both short-term and long-term prediction tasks within the dataset. Experimental results indicate that T-NIG is proficient in forecasting disease progression while maintaining disease-related characteristics, even when faced with an irregular temporal data distribution.
\end{abstract}


\section{Introduction}
\label{sec:introduction}
\IEEEPARstart{E}{}arly identification of disease presents an opportunity to decelerate its progression \cite{99}. Multiple studies have used diagnostic data for the purpose of prognosis \cite{84,85}, while others have used brain imaging techniques \cite{73,74,ref3} to anticipate the onset of the disease. The progression of deep learning technologies has facilitated the application of medical imaging generation methods \cite{ref4,ref5,ref7,ref8}, thus improving the prospects for the early detection of disease by image. 

 Image generation has the potential to provide medical professionals with a foundational imaging diagnosis for the prediction of AD. Inconsistent time intervals between sequential subject examinations present numerous challenges to imaging generation technologies. Research \cite{ref3,ref4,ref5,ref7,ref8} has sought to address this problem by synthesizing brain images for absent time points, thus improving the temporal consistency of brain imaging sequences of subjects. Although this strategy partially alleviates the issue of temporal inconsistency, the brain images generated, which attempt to estimate missing data, inevitably contain inaccuracies related to the disease deformation attributes of the images \cite{101}. Other researchers\cite{ref42,ref39,ref40,ref41,ref43} directly use brain imaging sequences with inconsistent time intervals for prediction. They  improve the accuracy of generating brain images in uneven time intervals by optimizing statistical methods, employing feature alignment techniques, iterative residual refinement, and designing weighted loss functions. 

The problem of inconsistent time intervals leads to a discrepancy between the extracted temporal features and the actual ones, resulting in a decline in the model's performance. Researchers \cite{ref7,ref66,ref2,102} have integrated the learning of disease image specificity with recurrent neural networks to facilitate the extraction of disease-related characteristics within the temporal domain. Performing long-term disease predictions is complicated by variable temporal intervals in image sequences. Existing methodologies continue to face challenges in preserving disease-related characteristics through extended image-generated predictions.

Previous research has struggled to preserve disease-related traits during the generation of long-sequence images at inconsistent time intervals. When applied as a prior distribution, the Normal Inverse Gamma (NIG) distribution produces a posterior distribution that remains within the same family \cite{zhang2019}. Recognizing that the timing parameters of the NIG distribution can reflect changes in disease-related features when images are not evenly spaced, this study addresses the challenge of long-term AD predictions when confronted with irregular time intervals in sequential images. 

This research presents a model for the generation of MRI images that applies the estimation of temporal parameters to the normal inverse gamma distribution (T-NIG). The main contributions of this study can be categorized into three essential areas.

(1) To address the challenge of generating MRI data with irregular time intervals, this paper introduces a generation method that uses the normal inverse gamma distribution for the temporal parameter (tNIG). This approach takes into account the timing parameters of the sequential image feature distribution.

(2) To preserve disease-related characteristics for long-term disease prediction, this article proposes a feature extraction module based on coordinate neighborhood graphs (TTCN and TDCN) that captures temporal changes in both texture and deformation characteristics.

(3) To address the challenge of the lack of temporal information, this paper proposes an algorithm that takes advantage of both aleatoric(AL) and epistemic uncertainty(EP) estimation to reduce errors caused by the lack of temporal information and errors in model prediction.

The subsequent sections of this manuscript are structured as follows. Section \ref{sec:relative} provides a review of the pertinent literature. In Section \ref{sec:method}, we delineate the proposed methodology. Section \ref{sec:experiment} outlines the experimental framework and presents the results obtained from the experiments conducted. Section \ref{sec:DISCUSSION} offers a discussion of the principal elements of the proposed method, along with an examination of the limitations inherent in the current study. Finally, we conclude the paper in Section \ref{sec:conclusion}.

\section{RELATED WORK}
\label{sec:relative}

This section discusses related work on long-term  disease prediction from three perspectives: Image prediction in inconsistent time intervals, preservation of disease-related features in brain imaging prediction, and reduction of high computational complexity.

\subsection{Image Prediction under Inconsistent Time Intervals}
The disease image data set is characterized by irregular intervals between samples from the imaging sequence, which presents a significant challenge for image generation techniques aimed at long-term  disease prediction. 
Recent studies \cite{ref7,ref42,ref39,ref41,ref43} have made substantial progress in this area. Pan \cite{ref7} developed a feature consistency generative network to ensure that the feature maps of the synthesized images align with those of their original counterparts, thus safeguarding information related to disease-related characteristics. However, these methodologies face challenges in consistently preserving disease-related characteristics during extended brain image prediction. Jung \cite{ref39} devised a temporal encoding mechanism coupled with recursive layers to improve prediction in the context of inconsistent time intervals. In addition, Li \cite{ref41} developed multifield transformations of all pairs, facilitating flexible feature extraction and alignment across varying time intervals. Furthermore, Ilyas \cite{ref43} integrated age-related information from subjects and used Wasserstein-based Generative Adversarial Networks (GAN) to generate brain images. However, the intrinsic instability of GAN training can cause issues such as mode collapse \cite{wgan}. 

Existing methods demonstrate notable strengths in handling irregular time intervals in preserving disease-related characteristics. However, these approaches face limitations, such as the GAN instability that leads to mode collapse and challenges in maintaining long-term feature consistency.
Considering the temporal transformation characteristics of feature distributions, the methodology proposed in this paper narrows the discrepancy between the generated images and the actual brain images across inconsistent time intervals.

\subsection{Image Features Maintaining in  Prediction}
Long-term prediction of brain images presents significant challenges in preserving disease-related characteristics. The NIG distribution \cite{ma2021,mead2015}, which is a key distribution in statistical analysis and Bayesian inference, can effectively capture these traits.
Recent studies have reported notable progress in this field \cite{ref70,ref66,ref46,ref47,ref69,ref40}. Salehi \cite{ref70} used a convolution neural network to extract temporal disease-related characteristics from brain images. Zhu \cite{ref40} used multiscale convolution layers to accurately capture temporal details related to disease on various scales within brain images, thus improving the retention of disease-related characteristics. 

The techniques evaluated show successful strategies for identifying disease-related characteristics, with convolutional neural networks particularly strong in extracting temporal features, and multiscale convolutions improving detail retention at various scales. However, these methods may face challenges in maintaining long-term feature consistency because they do not adequately model temporal dependencies close to important brain areas. Our proposed method addresses this issue by taking into account the temporal changes in disease-related characteristics near the brain image coordinates, which enhances the retention of these features in brain imaging prediction tasks.

\subsection{Uncertainty Estimation of NIG }
NIG distribution, when used as a prior distribution, yields a posterior distribution that remains within the same distribution\cite{zhang2019}.
 The distribution of NIG has attracted the interest of researchers\cite{zhang2017bayes,brown2010inference,goncu2016}. Ahmet\cite{goncu2016} found that financial data often exhibit characteristics such as leptokurtosis, fat-tailed, asymmetry, and volatility clustering.
In many real-world scenarios, the data often follow a normal distribution, but the true values of the mean and variance are unknown. The NIG distribution serves as a conjugate prior to the normal distribution when both the mean and the variance are unknown\cite{liu2022}. This conjugacy property is of great importance in Bayesian statistics. 

Due to the absence of temporal data in the prediction of brain images, researches \cite{ref26,78,ref31,ref32,ref33} have achieved advances in uncertainty estimation. 
Moz \cite{ref32} incorporated an inhibition factor into the Softmax function, enhancing the sensitivity of the model to the uncertainty of the data. However, this method requires comprehensive optimization strategies to determine an appropriate inhibition factor, which increases computational costs. Liu \cite{ref31} used distance calculations and weighted information to estimate uncertainty, thus reducing estimation errors through distance constraints. Kendall\cite{ref26} presents a framework that combines input-dependent aleatoric uncertainty with epistemic uncertainty and derives new loss functions from the uncertainty formulation. 

Techniques have been discussed that provide different strategies for estimating uncertainty. Moz's inhibition factor enhances sensitivity but necessitates expensive optimization, whereas Liu's distance-based constraints minimize errors but might not be widely applicable. Kendall's framework successfully integrates aleatoric and epistemic uncertainty, yet it does not focus on gaps in temporal data. This paper's approach creates specific forms of epistemic and aleatoric uncertainty, targeting errors caused by insufficient temporal information and those that follow a normal inverse gamma distribution.

\section{METHODS}
\label{sec:method}
\begin{figure*}[htbp]
\centering
\centerline{\includegraphics[width=2.12\columnwidth]{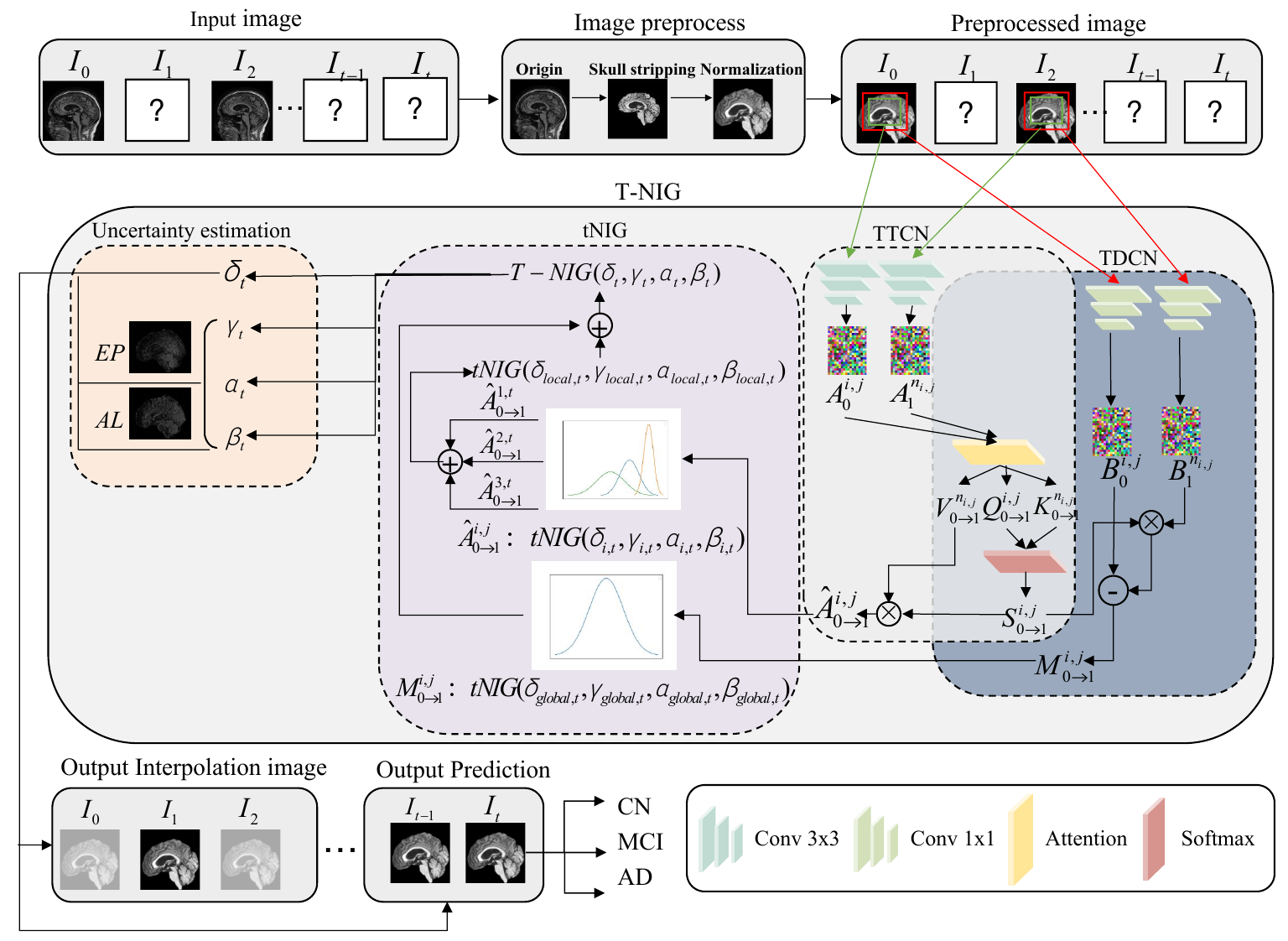}}
\caption{The T-NIG model employs brain images
from two different time points to create intermediate brain
images, forecast future images, and predict the disease. The original MRI scans $I_{0}$ and $I_{2}$ were segmented to remove the skull, followed by normalization. T-NIG extracts disease-related features from two preprocessed brain MRI images $I_{0}$ and $I_{2}$, using the TTCN and TDCN modules. The extracted information is then fed into the tNIG module for parameter fusion and parameter uncertainty estimation, thereby enabling the prediction and generation of brain images $I_{1}$,$I_{t-1}$, $I_{t}$.}
\label{Figure_1}
\end{figure*}

Long-term prediction of disease presents numerous challenges, including irregular time intervals between brain imaging sequences, difficulties in extracting disease-related features, and the computational complexity associated with large model parameters. To address these issues, the present study introduces a novel methodology known as the T-NIG general framework, as illustrated in Figure \ref{Figure_1}. Original MRI scans, designated as $I_{0}$ and $I_{2}$, underwent segmentation to remove the skull and were later normalized. The T-NIG framework extracts disease-related characteristics from the two preprocessed brain magnetic resonance images, $I_{0}$ and $I_{2}$, using the TTCN and TDCN modules. The information obtained is then processed through the tNIG module for parameter fusion and uncertainty estimation, thus enhancing the prediction and generation of brain images $I_{1}$,$I_{t-1}$, and $I_{t}$.

The Temporal parameter-Based Normal Inverse Gamma (T-NIG) algorithm for brain imaging prediction is comprised of three principal components: the extraction of Temporal Texture Features utilizing Coordinate Neighborhood (TTCN), the extraction of Temporal Deformation Features based on Coordinate Neighborhood (TDCN). The information obtained from these components is subsequently integrated through the fusion of the temporal parameters derived from the temporal Normal Inverse Gamma(tNIG). In addition, an uncertainty estimation module is employed to mitigate the observed discrepancies between predicted images and ground-truth images, which arise from insufficient temporal information.

\subsection{Definitions}
The objective of this paper is to construct the model $\mathcal{O} $, as shown in Equation\eqref{eq1}. Our goal is to generate brain images $\hat{I_{t}}\in \mathbb{R} ^{H\times W  } $ at any arbitrary time step $t$ from the brain images $I_{0},I_{1} \in \mathbb{R}^{H \times W }$. Here, $H, W$ represent the height and width of the brain images.
\begin{equation}\hat{I_{t}}= \mathcal{O} (I_{0}, I_{1},t).\label{eq1}\end{equation}

In this paper, we define the texture features of image $I_{0}$ as $A_{0}$ and those of image $I_{1}$ as $A_{1}$. The texture features of the regions in $A_{0} $ and $A_{1}$ are denoted as $A_{0}^{i,j} $ and $A_{1}^{i,j}$. $A_{1}^{n_{i,j} }$ represents the texture characteristics of the neighborhood around the region corresponding to $A_{0}^{i,j} $ in $A_{1}$. The dimensions of $A_{0} $,$A_{1}$,$A_{0}^{i,j}$,$A_{1}^{i,j}\in \mathbb{R} ^{\hat{H}\times \hat{W} \times C} $, while $A_{1}^{n_{i,j}}  \in \mathbb{R} ^{\hat{H}\times \hat{W}  \times \hat{C}}$. Here, $\hat{H},\hat{W}$ represent the height and width of the texture features. $C$ denotes the number of channels of texture features. $\hat{C}$ represents the number of neighborhood texture feature channels. The parameter $n$ indicates the size of the neighborhood window.

NIG is a distribution mixture algorithm, following the Normal Inverse Gamma distribution, as shown in Algorithm\ref{NIG}. NIG combines parameters of two different distributions $NIG(\delta _{1} ,\gamma _{1} ,\alpha _{1} ,\beta_{1} )$, $NIG(\delta _{2} ,\gamma _{2} ,\alpha _{2} ,\beta_{2} )$, as a mixture distribution $NIG(\delta ,\gamma ,\alpha ,\beta)$.

\begin{algorithm}[]
 \SetKwData{Left}{left}\SetKwData{This}{this}\SetKwData{Up}{up}
 \SetKwFunction{Union}{Union}\SetKwFunction{FindCompress}{FindCompress}
 \SetKwInOut{Input}{Input}\SetKwInOut{Output}{Output}
 \Input { NIG distribution\\
 $NIG(\delta _{1} ,\gamma _{1} ,\alpha _{1} ,\beta_{1} )$, $NIG(\delta _{2} ,\gamma _{2} ,\alpha _{2} ,\beta_{2} )$}
 \Output  {Mixture of NIG distribution: $NIG(\delta ,\gamma ,\alpha ,\beta)$}
 
 \BlankLine
 
  Calculate the mean parameter $\delta$ of the mixture distribution: $\delta =(\gamma _{1} +\gamma _{2} )^{-1} (\gamma _{1} \delta _{1}+\gamma _{2} \delta _{2} )$\;
  
  Calculate the scale parameter $\gamma$ of the mixture distribution: $\gamma =\gamma _{1} +\gamma _{2} $\;
  
  Calculate the shape parameter $\alpha$ of the mixture distribution: $\alpha =\alpha _{1} +\alpha _{2} +\frac{1}{2}$\;
 
  Calculate the shape parameter $\beta$ of the mixture distribution: $\beta =\beta _{1}+\beta _{2}+\frac{1}{2}   \gamma _{1} (\delta _{1}-\delta  )^{2} +\frac{1}{2} \gamma _{1} (\delta _{2}-\delta  )^{2}$\;
 
Output the resulting NIG distribution: $NIG(\delta ,\gamma ,\alpha ,\beta)$ \;
 \caption{NIG Distribution Mixture Algorithm}\label{NIG}
\end{algorithm}

The equation of NIG is shown in Equation\eqref{eq32}, $\alpha$ and $\beta$ are the shape and scale parameters. $\delta$ is the prior mean of the mean and $\gamma$ is the prior precision of the mean. $\Gamma (\cdot )$ represents the gamma function,  where $\delta, \gamma,\alpha,\beta\in \mathbb{R}$ with $\gamma >0,\alpha >1$,and $\beta >0$. Here, local parameters are denoted as $\delta_{global}$, $\gamma_{global}$ ,$\alpha_{global}$ ,$\beta _{global}$ and global parameters are $\delta_{global},\gamma_{global},\alpha_{global},\beta _{global}$. $\mu$ represents the mean of the normal distribution, and $\sigma ^{2}$ is the variance.

\begin{equation} 
\begin{split}
NIG(\delta,\gamma,\alpha,\beta)=&\sqrt{\frac{\gamma }{2\pi \sigma ^{2} } }\times \frac{\beta ^{\alpha } }{\Gamma (\alpha )} \times (\sigma ^{2} )^{-(\alpha +1)}\\&\times e^{(-\frac{\gamma (\mu -\delta )^{2} }{2\sigma ^{2} }-\frac{\beta }{\alpha }  )}.\label{eq32} 
\end{split}
\end{equation}

We design a distribution of tNIG (temporal parameter based on NIG)  to fit the distribution of changes in brain imaging features, as shown in Equation\eqref{eqtnig}.$\alpha_{t}$ and $\beta_{t}$ are the shape and scale parameters that incorporate temporal parameters. $\delta_{t}$ is the prior mean that includes temporal parameters, $\gamma_{t}$ is the prior precision of the mean with temporal parameters. 

\begin{equation} 
\begin{split}
tNIG(\delta_{t} ,\gamma_{t} ,\alpha_{t} ,\beta_{t} )=&\sqrt{\frac{\gamma_{t}  }{2\pi \sigma_{t}^{2}  } }\times \frac{\beta_{t}^{\alpha _{t} }  }{\Gamma (\alpha_{t}  )} \times (\sigma_{t}^{2}  )^{-(\alpha_{t}  +1)}\\&\times e^{(-\frac{\gamma_{t}  (\mu_{t}  -\delta )^{2} }{2\sigma_{t}^{2}  }-\frac{\beta_{t}  }{\alpha_{t}  }  )}.\label{eqtnig} 
\end{split}
\end{equation}

Since the mean and variance $(\mu,\sigma ^{2} )$ of the normal distribution are unknown, to model the distribution, we assume that $\mu$ and $\sigma^{2}$ follow a normal distribution and an inverse gamma distribution, as shown in Equation\eqref{eq14}, Equation\eqref{eq15}, and Equation\eqref{eq16}. Here, $d$ represents the variation in features, $\mathcal{N}$ denotes the normal distribution, and $ \Gamma^{-1}  (\cdot )$ represents the inverse gamma function.

\begin{equation}d\sim \mathcal{N} (\mu ,\sigma ^{2} ).\label{eq14}\end{equation}
\begin{equation}\mu \sim \mathcal{N} (\delta  ,\sigma ^{2}\gamma ^{-1}  ).\label{eq15}\end{equation}
\begin{equation}\sigma ^{2}  \sim \Gamma ^{-1} (\alpha ,\beta ).\label{eq16}\end{equation}

\subsection{Temporal Texture Feature Extraction from  Coordinate Neighborhood}

 As illustrated in Algorithm \ref{TTCN}, the TTCN module captures texture characteristics $A_{0}^{i,j}$,$A_{1}^{n_{i,j}}$ from images $I_{0}$ and $ I_{1}$, computing attention scores through query$Q_{0\to 1}^{i,j}$, key$K_{0\to 1}^{n_{i,j} }$ and $valueV_{0\to 1}^{n_{i,j} }$; then aggregating texture changes$\hat{A}_{0 \to 1}^{i,j}$. 

\begin{algorithm}[htp]
 \SetKwData{Left}{left}\SetKwData{This}{this}\SetKwData{Up}{up}
 \SetKwFunction{Union}{Union}\SetKwFunction{FindCompress}{FindCompress}
 \SetKwInOut{Input}{Input}\SetKwInOut{Output}{Output}
 
 \Input {Images sequence $I_{0}$, $ I_{1}$}
 \Output  {Texture change characteristics $\hat{A}_{0 \to 1}^{i,j}$} 

 \BlankLine
 Calculate the texture features of regions\\
\nonl $A_{0}^{i,j}$,$A_{1}^{n_{i,j}}$$\leftarrow$ Convolution($I_{0}, I_{1}$) \;
    
Compute the query matrix  \\
\nonl $Q_{0\to 1}^{i,j} \leftarrow A_{0}^{i,j} W_{q}$\;
  
Compute the key matrix  \\
\nonl  $K_{0\to 1}^{n_{i,j} } \leftarrow A_{1}^{n_{i,j} } W_{k}$\; 
  
Compute the value matrix \\
\nonl  $V_{0\to 1}^{n_{i,j} } \leftarrow A_{1}^{n_{i,j} } W_{v}$ \;

Compute the attention score matrix\\ 
 \nonl$S_{0\to 1}^{i,j}\leftarrow$ Softmax$(Q_{0\to 1}^{i,j}(K_{0\to 1}^{n_{i,j} } )^{T}/\sqrt{\hat{C} })$\;
  
Compute the texture change information \\ 
 \nonl $\hat{A}_{0 \to 1}^{i,j} \leftarrow S_{0\to 1}^{i,j}  V_{0\to 1}^{n_{i,j} }$ 
  \nonl $\hat{A}  _{0 \to 1}^{i,j}\sim tNIG(\delta_{k}, \gamma_{k}, \alpha_{k}, \beta_{k} )$, $k\in {1,2,3}$\;
 \caption{TTCN}
 \label{TTCN}
\end{algorithm}

First, TTCN obtains the texture characteristics $A_{0}$ and $A_{1}^{n_{i,j}}$ from two sequential brain images $I_{0}$ and $ I_{1}$, to calculate the enhanced characteristics of the texture change $\hat{A}_{0 \to 1}^{i,j}$. 

Second, TTCN generate the query $Q_{0\to 1}^{i,j}$, the key/value $K_{0\to 1}^{n_{i,j} }$ and $V_{0\to 1}^{n_{i,j} }$ through the weight coefficients $W_{q}, W_{k}, W_{v} $, where $W_{q}, W_{k}, W_{v}\in \mathbb{R} ^{C\times \hat{C} }$.This process is shown in Equation\eqref{eq2}, Equation\eqref{eq3} and Equation\eqref{eq4}.
\begin{equation}Q_{0\to 1}^{i,j} =A_{0}^{i,j} W_{q},\label{eq2}\end{equation}
\begin{equation}K_{0\to 1}^{n_{i,j} } =A_{1}^{n_{i,j} } W_{k},\label{eq3}\end{equation}
\begin{equation}V_{0\to 1}^{n_{i,j} } =A_{1}^{n_{i,j} } W_{v},\label{eq4}\end{equation}

Third, the attention map $S_{0\to 1}^{i,j}$ is calculated to extract the temporal changes in texture, as shown in Equation\eqref{eq5}. The attention map $S_{0\to 1}^{i,j} \in \mathbb{R} ^{n\times n} $ is obtained by the attention mechanism, which calculates the dot product between the query $Q_{0\to 1}^{i,j}$ and the key/value $K_{0\to 1}^{n_{i,j} }$ at each position. Here, $T$ denotes the matrix transpose and $\hat{C}$ represents the number of channels. 
\begin{equation} S_{0\to1}^{i,j}= Softmax(Q_{0\to 1}^{i,j}(K_{0\to 1}^{n_{i,j} } )^{T}/\sqrt{\hat{C} }   ).\label{eq5}\end{equation}

Fourth, the texture change information $\hat{A} _{0 \to 1}^{i,j}$ for the corresponding regions in $I_{0}$ and $I_{1}$ is calculated through $S_{0\to 1}^{i,j}  V_{0\to 1}^{n_{i,j} }$, as shown in Equation\eqref{eq6}.
\begin{equation}\hat{A}_{0 \to 1}^{i,j} = S_{0\to 1}^{i,j}  V_{0\to 1}^{n_{i,j} } .\label{eq6}\end{equation}

Finally, $\hat{A} _{0 \to 1}^{i,j}$ encapsulates the changes in texture within the corresponding regions of the two images of the brain. In this paper, $k\in {1,2,3}$ represents different scales of temporal texture feature maps generated by 1×1, 3×3, and 5×5 convolution kernels. This approach facilitates the transformation of texture features between images. Furthermore, $\hat{A}_{0 \to 1}^{i,j}$ follows the tNIG distribution, as demonstrated in Equation \eqref{eq61}.
\begin{equation}
\hat{A}  _{0 \to 1}^{i,j}\sim tNIG(\delta_{k,t}, \gamma_{k,t}, \alpha_{k,t}, \beta_{k,t} ), k\in {1,2,3}.\label{eq61}
\end{equation}

\subsection{Temporal Deformation Feature Extraction from  Coordinate Neighborhood}

Taking into account the localized atrophy observed in brain images, the TDCN module, as illustrated in Algorithm \ref{TDCN}, extracts the features of temporal deformation $M_{0\to 1}^{i,j}$ by first building neighborhood maps from $B_{0}^{i,j} $ and $B_{1}^{n_{i,j}}$. Then it estimates the deformation features \( M_{0\to 1}^{i,j} \) through attention-based coordinate shifts. This extraction process helps constrain changes in the disease-related features of the predicted images.

\begin{algorithm}[htp]
 \SetKwData{Left}{left}\SetKwData{This}{this}\SetKwData{Up}{up}
 \SetKwFunction{Union}{Union}\SetKwFunction{FindCompress}{FindCompress}
 \SetKwInOut{Input}{Input}\SetKwInOut{Output}{Output}
 \Input {Images sequence $I_{0}$, $ I_{1}$}
 \Output  {Temporal deformation feature: $M_{0\to 1}^{i,j}$}
 \BlankLine
 
Calculate the neighborhood of $B_{0}^{i,j} $ and $B_{1}^{n_{i,j}}$\\
\nonl $B_{0}^{i,j}$, $B_{1}^{n_{i,j}}$$\leftarrow$Convolution($I_{0} , I_{1}$)\;
    
Compute the query matrix  \\
\nonl   $Q_{0\to 1}^{i,j} \leftarrow A_{0}^{i,j} W_{q}$\;
  
Compute the key matrix \\
\nonl  $K_{0\to 1}^{n_{i,j} } \leftarrow A_{1}^{n_{i,j} } W_{k}$\; 

Compute the value matrix \\
\nonl  $V_{0\to 1}^{n_{i,j} } \leftarrow A_{1}^{n_{i,j} } W_{v}$ \;
  
Compute the attention score matrix \\
\nonl  $S_{0\to 1}^{i,j}\leftarrow Softmax (Q_{0\to 1}^{i,j}(K_{0\to 1}^{n_{i,j} } )^{T}/\sqrt{\hat{C} })$\;

Compute the global temporal deformation estimation 
 
 \nonl $M_{0\to 1}^{i,j} \leftarrow S_{0\to 1}^{i,j} B_{1}^{n_{i,j} } -B_{0}^{i,j}$ 
 
\nonl $M_{0\to 1}^{i,j}\sim tNIG(\delta_{global} ,\gamma_{global},\alpha_{global} ,\beta_{global} )$\;

\caption{TDCN}
\label{TDCN}
\end{algorithm}

First, TDCN constructs neighborhood maps $B\in \mathbb{R} ^{\hat{H}\times \hat{W} } $for brain images $I_{0}$,$I_{1}$. $B_{0}^{i,j}$ represents the neighborhood of $A_{0}^{i,j} $. $B_{1}^{n_{i,j} } $ represents the neighborhood of $A_{1}^{i,j} $.

Second,  TDCN generate the query $Q_{0\to 1}^{i,j}$, the key/value $K_{0\to 1}^{n_{i,j} }$ and $V_{0\to 1}^{n_{i,j} }$ through the weight coefficients $W_{q}, W_{k}, W_{v} $. The deformation vectors are calculated based on the neighborhood maps. The temporal deformation vector $M_{0\to 1}^{i,j}$, as shown in Equation\eqref{eq7}, is obtained by subtracting its original position $B_{0}^{i,j}$ from its corresponding position $S_{0\to 1}^{i,j} B_{1}^{n_{i,j} }$. Using the attention map $S_{0\to 1}^{i,j}$ to weight the coordinates of the neighborhood, the corresponding position of $A_{1}^{i,j} $  in the brain image $I_{1}$ is estimated as $S_{0\to 1}^{i,j} B_{1}^{n_{i,j} }$. 

\begin{equation}M_{0\to 1}^{i,j} =S_{0\to 1}^{i,j} B_{1}^{n_{i,j} } -B_{0}^{i,j}  .\label{eq7}\end{equation}

$M_{0\to 1}^{i,j}$ contains temporal deformation information , as shown in Equation\eqref{eq33}, following the tNIG distribution with global parameter, which provides prior knowledge for the estimation of temporal deformations.

\begin{equation}
    M_{0\to 1}^{i,j}\sim tNIG(\delta_{global, t} ,\gamma_{global, t},\alpha_{global, t} ,\beta_{global, t} ).\label{eq33}
\end{equation}

\subsection{Uncertainty Estimation under Uneven Distribution of Temporal Image }

Due to the absence of temporal information in the datasets, errors associated with disease-related features and inherent stochastic variations are evident in long-term image predictions. These factors contribute to both aleatoric (AL) and epistemic (EP) uncertainties. AL refers to the discrepancies between the model predictions and the actual values, while EP arises from insufficient temporal data.

The parameter uncertainty estimation of T-NIG aims to evaluate both AL and EP, thus constraining and mitigating the adverse effects of uncertainty on T-NIG and ultimately enhancing the quality of predicted brain images. The characteristic change is indicated as $d$, as expressed in Equation\eqref{eq17}, AL is shown in Equation\eqref{eq18}, and EP is presented in Equation\eqref{eq19}, where $t$ represents the temporal parameter, $\mathbb{E}$ denotes the expectation, and $Var$ denotes the variance.

\begin{equation}d=\mathbb{E} (\mu_{t} )=\delta_{t} .\label{eq17}\end{equation}
\begin{equation}AL=\mathbb{E}(\sigma_{t}^{2} ) =\frac{\beta_{t} }{\alpha_{t} -1} .\label{eq18}\end{equation}
\begin{equation}EP=Var(\mu_{t} )=\frac{\beta_{t} }{\gamma_{t} (\alpha_{t} -1)} .\label{eq19}\end{equation}

\subsection{Loss Function for Long-term Image Prediction}

The total loss function $\mathcal{L}$ of T-NIG is given by Equation\eqref{eq31}. $\mathcal{L} _{rec}$ is utilized to evaluate both the pixel-wise discrepancies between predicted and real images , whereas $\mathcal{L}^{U}(w)$ quantifies the confidence levels of the model's predictions. 

\begin{equation}\mathcal{L} =\mathcal{L} _{rec} +  \mathcal{L}^{U}(w).\label{eq31}\end{equation}

T-NIG uses the reconstruction loss $\mathcal{L}_{rec}$ to compute the loss between the generated image and the real image, where $\hat{I}_{t}$ represents the predicted image, $I_{t}$ is the real image, and $f$ denotes the pixel loss, as shown in Equation\eqref{eq27}.

\begin{equation}\mathcal{L}_{rec} =f(\hat{I}_{t} ,I_{t}  ).\label{eq27}\end{equation}

To improve the generalization capability of the model and prevent overfitting, T-NIG has designed the loss function $\mathcal{L}^{U}(w)$ by $\mathcal{L}_{i}^{N} (w) $ and $\mathcal{L}_{i}^{R} (w)$. $w$ represents the set of all parameters. When $\mathcal{L}_{i}^{N} (w) $ is used to assess the confidence of the T-NIG parameter and when $\mathcal{L} ^{R} (w)$ is used to assess the confidence of the changes in characteristics. $\mathcal{L}_{i}^{R} (w)$  is defined as the expectation over all pixels, where $\tau>0$ controls the regularization strength, $N$ is the total number of pixels, and $i$ represents the i-th image, as shown in Equation\eqref{eq30}.

\begin{equation}\mathcal{L}^{U}(w)=\frac{1}{N}   \sum_{i=0}^{N-1} (\mathcal{L}_{i}^{N} (w)+\tau \mathcal{L}_{i}^{R}(w) ).\label{eq30}\end{equation}

The loss function $\mathcal{L} ^{N} (w)$ measures the confidence predictions of the T-NIG to reduce the error between the predicted values and the true values. During the training process, this function is defined as the negative Log of the T-NIG parameters, with \( \Gamma \) denoting the gamma function, as shown in Equation\eqref{eq28}. Here, \( \Omega = 2\gamma + \alpha \), where \( w \) represents the collection of estimated distribution parameters.

\begin{equation}
\begin{split}
    \mathcal{L} ^{N} (w)=&\frac{1}{2} log(\frac{\pi }{\gamma } )-\alpha log(\Omega )+log(\frac{\Gamma (\alpha )}{\Gamma (\alpha +\frac{1}{2} )} ) .\label{eq28}
\end{split}
\end{equation}

The function $\mathcal{L}^{R} (w)$ is used to minimize inaccuracies in the predictions, where $d$ represents the ground truth of the feature change map, as illustrated in Equation\eqref{eq29}.

\begin{equation}\mathcal{L} ^{R} (w)=\mid d -\delta \mid \cdot(2\gamma +\alpha ).\label{eq29}\end{equation}

\subsection{Temporal Parameter Estimation of Disease-related Feature Distribution}

The tNIG function presented in this paper adheres to a normal distribution.
According to the central limit theorem, the sum of a large number of independent and identically distributed random variables tends to approximate a normal distribution. The inverse gamma distribution is frequently used to model the variance of a normal distribution.
In Bayesian statistics, the NIG distribution serves as a conjugate prior distribution for the parameters of a normal distribution. When using Bayesian methods to estimate the parameters of a normal distribution, selecting the NIG distribution as the prior results in a posterior distribution that is also a NIG distribution \cite{nig}.

In this paper, the parameters of the NIG distribution are derived through parameter estimation. As shown in Algorithm \ref{T-NIG}, the algorithm predicts the image $\hat{I_{t}}$ by extracting the temporal texture and deformation characteristics through TTCN and TDCN. Subsequently, it fuses local and global tNIG distributions to compute the Aleatoricn uncertainty and Epistemic uncertainty. Finally, the predicted image $\hat{I_{t}}$ is generated by deconvolution.

T-NIG is obtained by combining tNIG, as shown in Equation\eqref{eq25}. Here, $\oplus$ represents the summation operation for the NIG distributions \cite{ref72}. 

\begin{equation}
\begin{split}
     &T-NIG(\delta_{t}  ,\gamma_{t}  ,\alpha_{t}  ,\beta_{t}  )= \\
  &tNIG(\delta _{local,t} ,\gamma _{local,t} ,\alpha _{local,t} ,\beta_{local,t} )\oplus \\ 
 &tNIG(\delta _{global,t} ,\gamma _{global,t} ,\alpha _{global,t} ,\beta_{global,t}).\label{eq25}
\end{split}
\end{equation}

The tNIG module is designed to capture the relationship between changes in the temporal distribution of disease-related characteristics. The local parameter fusion tNIG module integrates three tNIG distributions into a single distribution, as shown in Equation\eqref{eq24}.

\begin{equation}
    \begin{split}
        &tNIG(\delta _{local,t} ,\gamma _{local,t} ,\alpha _{local,t} ,\beta _{local,t} )=\\
 &tNIG(\delta _{1,t} ,\gamma _{1,t} ,\alpha _{1,t} ,\beta_{1,t} )\oplus  
 tNIG(\delta _{2,t} ,\gamma _{2,t} ,\alpha _{2,t} ,\beta_{2,t} )\oplus \\
 &tNIG(\delta _{3,t} ,\gamma _{3,t} ,\alpha _{3,t} ,\beta_{3,t} )  .\label{eq24}
    \end{split}
\end{equation}

\begin{algorithm}[htbp]
 \SetKwData{Left}{left}\SetKwData{This}{this}\SetKwData{Up}{up}
 \SetKwFunction{Union}{Union}\SetKwFunction{FindCompress}{FindCompress}
 \SetKwInOut{Input}{Input}\SetKwInOut{Output}{Output}
 
 \Input {Images sequence $I_{0}$, $ I_{1}$}
 \Output  {Predicted image $\hat{I_{t}}$} 
 
Calculate the temporal texture feature\\ 
\nonl $\hat{A}_{0 \to 1}^{i,j} \leftarrow$ TTCN($I_{0} , I_{1}$) \;
 
Calculate the temporal deformation feature 
$M_{0\to 1}^{i,j} \leftarrow $TDCN($I_{0} , I_{1}$) \;

Calculate NIG local distributions
$tNIG(\delta _{local,t} ,\gamma _{local,t} ,\alpha _{local,t} ,\beta _{local,t})\leftarrow tNIG(\delta _{1,t} ,\gamma _{1,t} ,\alpha _{1,t} ,\beta_{1,t} )\oplus tNIG(\delta _{2,t} ,\gamma _{2,t} ,\alpha _{2,t} ,\beta_{2,t} )\oplus tNIG(\delta _{3,t} ,\gamma _{3,t} ,\alpha _{3,t} ,\beta_{3,t} )$ \;

 Summation  local and global distributions
 $T-NIG(\delta_{t}  ,\gamma_{t}  ,\alpha_{t}  ,\beta_{t}  ) \leftarrow tNIG(\delta _{local,t} ,\gamma _{local,t} ,\alpha _{local,t} ,\beta_{local,t} )\oplus tNIG(\delta _{global,t} ,\gamma _{global,t} ,\alpha _{global,t} ,\beta_{global,t})$ \;

\textit{Using T-NIG  parameters to compute Aleatoricn uncertainty and Epistemic uncertainty}\\
\nonl Calculate feature change $d \leftarrow \delta_{t}$ \;
\nonl  Calculate Aleatoricn uncertainty $AL \leftarrow \frac{\beta_{t} }{\alpha_{t} -1}$\;
\nonl  Calculate Epistemic uncertainty $EP \leftarrow \frac{\beta_{t} }{\gamma_{t} (\alpha_{t} -1)}$ \;
 
 Predicted image:
 $\hat{I_{t}} \leftarrow$ DeConvolution(d, AL, EP)\;
 \caption{T-NIG}
 \label{T-NIG}
\end{algorithm}

\section{EXPERIMENTS}
\label{sec:experiment}
The T-NIG model uses brain images from two time points to generate intermediate brain images , generate future brain images, and realize the long-term AD predicted.

To validate the performance of T-NIG, we conduct several experiments in this section. First, we compare various methods for both short- and long-term AD prediction. In addition, we evaluate the quality of image generation in long-term prediction tasks to verify the retention of disease-related features in the generated images. Finally, we performed ablation studies on each module to assess their effectiveness within T-NIG.

\subsection{Experimental Settings}

The data sets used in this study include ADNI-1 and ADNI-2\footnote{ADNI official website: http://adni.loni.usc.edu \label{FOOT1}}. The data set has undergone an ethical review to facilitate data sharing with all interested investigators. The ADNI data sets integrate brain imaging data from multiple research centers, encompassing various groups, including normal controls (CN), mild cognitive impairment (MCI) and AD. ADNI-1 was launched in October 2005, with data collection continuing until approximately 2010. ADNI-2 began in 2011, building on and expanding previous research directions. Data collection for ADNI-2 continued until 2017. The irregularity in data collection leads to unevenly spaced time intervals in brain image. The distribution of classes, the consistent and inonsistent sequence in ADNI-1 and ADNI-2 are presented in Table \ref{Table1}.

\begin{table}[htbp]
\centering
\caption{The distribution of classes, the consistent and inonsistent sequence in ADNI-1 and ADNI-2.}
\begin{tabular}{c|p{25pt}p{25pt}p{25pt}|cc}
\hline
\multirow{3}{*}{Dataset} & \multicolumn{3}{c|}{Class} & \multicolumn{2}{c}{Sequential Time Interval}       \\ \cline{2-6} 
                         & CN      & MCI     & AD     & Consistent & Inonsistent\\
                                \hline
ADNI-1                   & 298     & 317     & 275    & 63              & 827              \\
ADNI-2                   & 276     & 328     & 271    & 75              & 820              \\ \hline
\end{tabular}
\label{Table1}
\end{table}

In this study, the quality of image generation is evaluated using SSIM, PSNR, and MSE. To evaluate classification performance, the Acc, Precision and F1 scores are utilized. In the tables presented, the symbol $\uparrow$ indicates that a higher value is better, while $\downarrow$ indicates that a lower value is better.

Both data sets underwent the same preprocessing pipeline. Initially, the original MRI scans were segmented using HD-BET \cite{120} to remove the skull, followed by correction of head orientation and normalization using FSL\cite{121}. The processed images were resized to $218\times182\times218$.

The training parameters used in this experiment include 200 epochs, an Adam learning rate of 0.0001, and weight decay regularization with a parameter setting of $1\times10^{-5} $. The batch size was set to 8 and a 5-fold cross-validation was used. The experiments were performed on an Nvidia RTX 3090 GPU.

\subsection{Performance Evaluates of Prediction and Classification}

This section evaluates the performance of various methods in both short-term and long-term brain image prediction and three-class classification tasks. It compares the quality of brain image sequences predicted for 5 and 10 years in the future and validates the performance of disease state predictions for the categories of CN, MCI, and AD.

\subsubsection{Evaluation of the brain image sequence prediction tasks}

Table \ref{Table2} presents an evaluation of the brain image sequence prediction tasks over 5- and 10-year periods.

In the 5-year brain image sequence prediction task, T-NIG achieved the best short-term prediction results in both the ADNI-1 and ADNI-2 datasets. For example, in the ADNI-1 dataset, the SSIM of T-NIG was 0.082 higher than that of DIS\cite{ref7}, which had the second-best performance. Similarly, in the ADNI-2 dataset, the SSIM of T-NIG was 0.092 higher than that of DIS, which also ranked second.

In the 10-year brain image sequence prediction task, T-NIG once again achieved the best long-term prediction results in the ADNI-1 and ADNI-2 datasets. For example, in the ADNI-1 dataset, the SSIM of T-NIG was 0.128 higher than that of DIS, which had the second-best performance. Similarly, in the ADNI-2 dataset, the SSIM of T-NIG was 0.137 higher than that of DIS, which again ranked second.

\begin{table}[htbp]
\centering
\caption{Brain Image Sequential Prediction (CN vs. MCI vs. AD).}
\setlength{\tabcolsep}{0.8mm}{} 
\begin{tabular}{c|cccccc}
\hline
\multirow{3}{*}{Method} & \multicolumn{6}{c}{5-year sequence}                                    \\ \cline{2-7} 
                        & \multicolumn{3}{c|}{ADNI-1}               & \multicolumn{3}{c}{ADNI-2} \\ \cline{2-7} 
                        & SSIM$\uparrow$  & PSNR$\uparrow$ & \multicolumn{1}{c|}{MSE$\downarrow$}   & SSIM$\uparrow$    & PSNR$\uparrow$   & MSE$\downarrow$     \\ \hline
DRM\cite{ref39}                     & 0.669 & 23.6 & \multicolumn{1}{c|}{1.314} & 0.648   & 21.0   & 1.712   \\
LRE\cite{ref40}                     & 0.693 & 24.2 & \multicolumn{1}{c|}{1.217} & 0.692   & 21.3   & 1.781   \\
AMT\cite{ref41}                     & 0.786 & 27.6 & \multicolumn{1}{c|}{0.894} & 0.753   & 26.3   & 0.237   \\
DIS\cite{ref7}                     & 0.877 & 31.9 & \multicolumn{1}{c|}{0.517} & 0.855   & 30.1   & 0.841   \\
WGAN\cite{ref43}                    & 0.854 & 31.4 & \multicolumn{1}{c|}{0.570} & 0.815   & 29.5   & 0.781   \\
T-NIG(Ours)             & $\mathbf{0.919}$ & $\mathbf{34.1}$ & \multicolumn{1}{c|}{$\mathbf{0.302}$} & $\mathbf{0.947}$   & $\mathbf{35.7}$   & $\mathbf{0.378 }$  \\ \hline
\multirow{3}{*}{Method} & \multicolumn{6}{c}{10-year sequence}                                   \\ \cline{2-7} 
                        & \multicolumn{3}{c|}{ADNI-1}               & \multicolumn{3}{c}{ADNI-2} \\ \cline{2-7} 
                        & SSIM$\uparrow$  & PSNR $\uparrow$& \multicolumn{1}{c|}{MSE$\downarrow$}   & SSIM$\uparrow$    & PSNR $\uparrow$  & MSE$\downarrow$     \\ \hline
DRM\cite{ref39}                     & 0.548 & 20.1 & \multicolumn{1}{c|}{1.856} & 0.517   & 17.5   & 3.687   \\
LRE\cite{ref40}                     & 0.616 & 21.3 & \multicolumn{1}{c|}{1.781} & 0.598   & 19.4   & 2.716   \\
AMT\cite{ref41}                     & 0.701 & 23.4 & \multicolumn{1}{c|}{0.916} & 0.671   & 24.3   & 1.798   \\
DIS\cite{ref7}                     & 0.795 & 30.4 & \multicolumn{1}{c|}{0.745} & 0.765   & 28.5   & 1.367   \\
WGAN\cite{ref43}                    & 0.782 & 29.7 & \multicolumn{1}{c|}{0.805} & 0.751   & 26.1   & 1.164   \\
T-NIG(Ours)             & $\mathbf{0.903}$ & $\mathbf{33.0}$ & \multicolumn{1}{c|}{$\mathbf{0.387}$} & $\mathbf{0.902}$   & $\mathbf{34.2}$   & $\mathbf{0.418}$   \\ \hline
\end{tabular}
\label{Table2}
\end{table}

Compared to the 5-year brain image sequence prediction task, the performance of all methods decreased to some extent in the 10-year brain image sequence prediction task, with T-NIG exhibiting the smallest performance decline. For example, in the ADNI-1 dataset, the SSIM for T-NIG decreased by 0.036, while the SSIM for DIS, which demonstrated the second best performance, decreased by 0.082. These experimental results indicate that T-NIG is robust in both short-sequence and long-sequence brain image prediction tasks.

\subsubsection{Evaluation of the brain image sequence classification tasks}

Table \ref{Table3} presents an evaluation of the three-class classification tasks for brain image sequences over 5-year and 10-year periods.

In the 5-year brain image sequence classification task, T-NIG achieved the highest performance in the ADNI-1 and ADNI-2 datasets. For example, the Acc of T-NIG was 0.067 higher than that of DIS in the ADNI-1 dataset, which ranked second in performance.

In the 10-year brain image sequence prediction task, T-NIG once again achieved the best results in the ADNI-1 and ADNI-2 datasets. For example, in the ADNI-1 dataset, the Acc of T-NIG was 0.089 higher than that of DIS, the second best performer. Similarly, in the ADNI-2 dataset, the Acc of T-NIG exceeded that of DIS by 0.091, with DIS ranking as the second best performer.

\begin{table}[htbp]
\centering
\caption{Brain Image Sequential Classification(CN vs. MCI vs. AD).}
\setlength{\tabcolsep}{1.2mm}{} 
\begin{tabular}{c|cccccc}
\hline
\multirow{3}{*}{Method} & \multicolumn{6}{c}{5-year sequence}                                    \\ \cline{2-7} 
                        & \multicolumn{3}{c|}{ADNI-1}               & \multicolumn{3}{c}{ADNI-2} \\ \cline{2-7} 
                        & Acc$\uparrow$  & Pre$\uparrow$ & \multicolumn{1}{c|}{F1$\uparrow$}   & Acc$\uparrow$    & Pre$\uparrow$   & F1$\uparrow$     \\ \hline
DRM\cite{ref39}                     & 0.679 & 0.672 & \multicolumn{1}{c|}{0.670} & 0.668   & 0.659   & 0.662   \\
LRE\cite{ref40}                     & 0.732 & 0.734 & \multicolumn{1}{c|}{0.725} & 0.725   & 0.712   & 0.716   \\
AMT\cite{ref41}                     & 0.838 & 0.835 & \multicolumn{1}{c|}{0.827} & 0.807   & 0.812   & 0.798   \\
DIS\cite{ref7}                     & 0.901 & 0.908 & \multicolumn{1}{c|}{0.897} & 0.874   & 0.879   & 0.867   \\
WGAN\cite{ref43}                    & 0.877 & 0.875 & \multicolumn{1}{c|}{0.861} & 0.873   & 0.851   & 0.841   \\
T-NIG(Ours)             & $\mathbf{0.948}$ & $\mathbf{0.941}$ & \multicolumn{1}{c|}{$\mathbf{0.937}$} & $\mathbf{0.931}$   & $\mathbf{0.928}$   & $\mathbf{0.919}$   \\ \hline
\multirow{3}{*}{Method} & \multicolumn{6}{c}{10-year sequence}                                   \\ \cline{2-7} 
                        & \multicolumn{3}{c|}{ADNI-1}               & \multicolumn{3}{c}{ADNI-2} \\ \cline{2-7} 
                        & Acc$\uparrow$  & Pre$\uparrow$ & \multicolumn{1}{c|}{F1$\uparrow$}   & Acc$\uparrow$    & Pre$\uparrow$   & F1$\uparrow$     \\ \hline
DRM\cite{ref39}                     & 0.624 & 0.631 & \multicolumn{1}{c|}{0.659} & 0.615   & 0.627   & 0.606   \\
LRE\cite{ref40}                     & 0.697 & 0.715 & \multicolumn{1}{c|}{0.680} & 0.673   & 0.681   & 0.665   \\
AMT\cite{ref41}                     & 0.764 & 0.784 & \multicolumn{1}{c|}{0.743} & 0.738   & 0.724   & 0.731   \\
DIS\cite{ref7}                     & 0.846 & 0.863 & \multicolumn{1}{c|}{0.837} & 0.826   & 0.831   & 0.818   \\
WGAN\cite{ref43}                    & 0.828 & 0.831 & \multicolumn{1}{c|}{0.794} & 0.804   & 0.798   & 0.782   \\
T-NIG(Ours)             & $\mathbf{0.915}$ & $\mathbf{0.912}$ & \multicolumn{1}{c|}{$\mathbf{0.907}$} & $\mathbf{0.897}$   & $\mathbf{0.884}$   & $\mathbf{0.891}$   \\ \hline
\end{tabular}
\label{Table3}
\end{table}

Compared to the 5-year brain image sequence classification task, the performance of all methods decreased to some extent in the 10-year brain image sequence classification task, with T-NIG experiencing the smallest decline. For example, in the ADNI-1 dataset, the Acc of T-NIG decreased by 0.033, while the Acc of DIS, the second-best performer, decreased by 0.055. These experimental results demonstrate that T-NIG is robust in both short- and long-sequence brain image classification tasks.

\subsection{Disease-related Feature Evaluation}

This section evaluates the ability to maintain disease-related characteristics in the prediction of brain images by comparing 10-year sequences of brain images produced by each method, along with an analysis of the predicted images.

\subsubsection{Brain images generation comparison of interpolation and prediction}

\begin{figure}[htbp]
\centering
\centerline{\includegraphics[width=1\columnwidth]{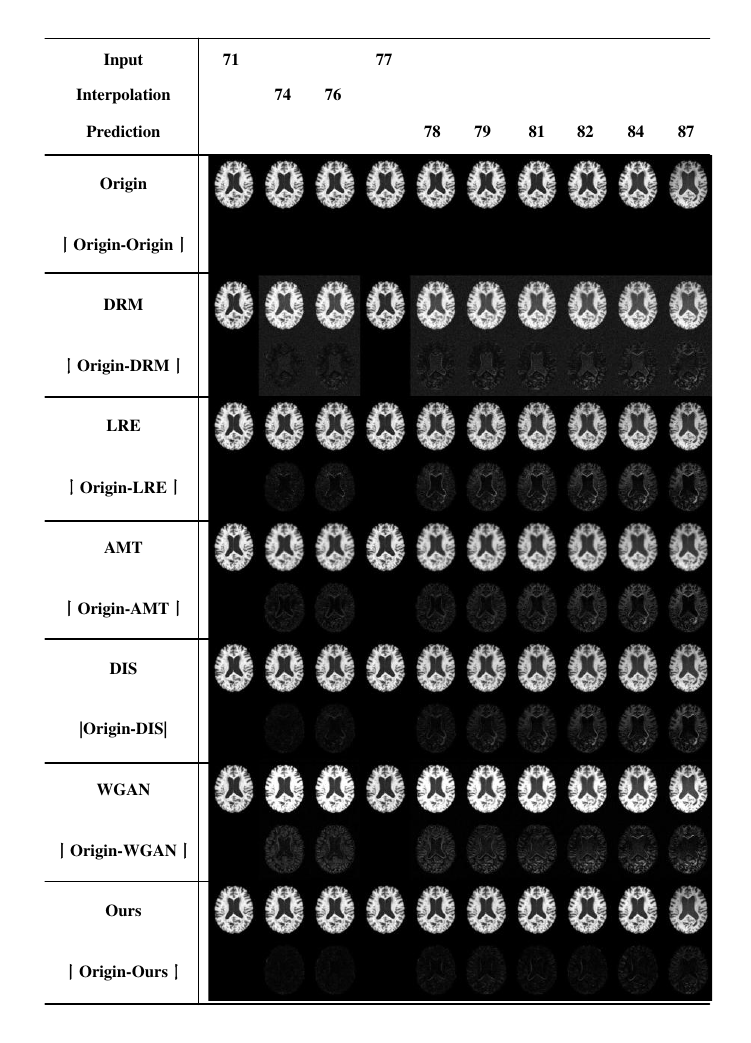}}
\caption{Long-term brain image generation in ADNI-1. Each model utilizes brain images of subjects as input at ages 71 and 77.The brain images for ages 74 and 76 are generated through interpolation, whereas the brain images for ages 78 to 87 are predictions. In each group of images, the first row displays the predicted brain images, while the second row shows the differences between the predicted and actual brain images. }
\label{Figure_2}
\end{figure}

Figure \ref{Figure_2} illustrates the interpolated and predicted brain images generated from the brain images of a subject at ages 71 and 77 in ADNI-1. The first row represents the ages of input, interpolation, and output. The subsequent rows display the brain images produced by the corresponding methods, along with the differences between the generated images and the original images. A higher number of white areas in the differential images indicates a greater discrepancy between the brain images generated and the originals. 

Interpolation of images at ages 74 and 76: The images produced by the DRM \cite{ref39} exhibit significant background noise. In contrast, the brain images generated by DIS and T-NIG show minimal differences from the original brain images.

Prediction of images from ages 78 to 87: There are notable discrepancies in the deformable features between the brain images predicted by LRE, AMT \cite{ref41}, DIS, and WGAN \cite{ref43} compared to the actual brain images. The 84-year-old brain images predicted by DRM display considerable background noise. In contrast, the deformable features of the 84-year-old brain images predicted by T-NIG align closely with those of the original brain images, demonstrating the smallest deviation from the actual brain images.

\subsubsection{Deformation detail comparison of brain images generation}

\begin{figure*}[htbp]
    \centering
    \subfloat[The interpolated of 74-year-old:The brain images of a subject at the ages of 71 and 77 are used for interpolation to generate the brain image at the age of 74.]{     \includegraphics[width=0.9\columnwidth]{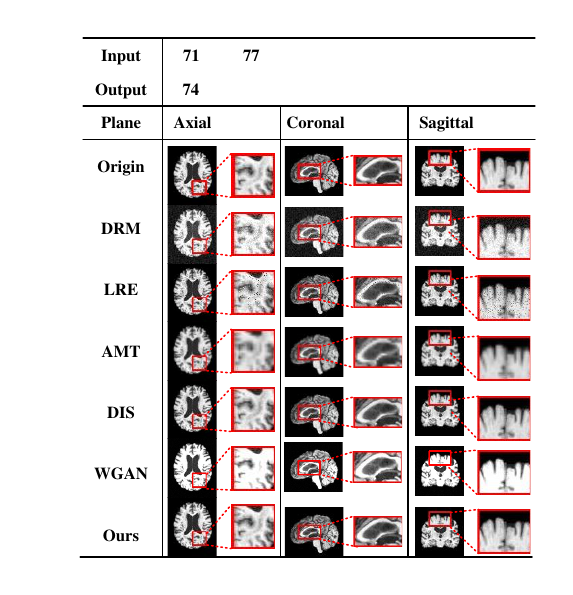}
        \label{Figure_3a}
    }
    \hfill
    \subfloat[The predicted image 74-year-old: The brain images of a subject at the ages of 71 and 77 are used for prediction to generate the brain image at the age of 84.]{      \includegraphics[width=0.9\columnwidth]{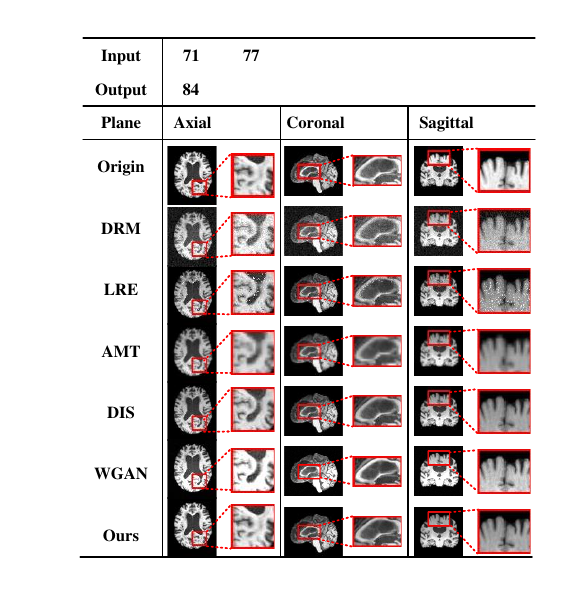}
        \label{Figure_3b}
    }
    \caption{A detailed comparison of the three planes of the interpolated brain images and the predicted brain images in ADNI-1 generated by various methods is presented. The input images are from a 71-year-old and a 77-year-old individual, and output images are 74 and 84. Axial, Coronal and Sagittal represent the three planes in brain image.}
    \label{Figure_3}
\end{figure*}

Figure \ref{Figure_3} presents a detailed comparison of brain images of a subject at ages 74 and 84, generated from images of the same subject at ages 71 and 77 in ADNI-1. 

Regarding texture characteristics, using the axial plane as an example, brain images of 74-year-olds and 84-year-olds generated by DRM exhibit white noise in both the brain and background regions. In contrast, the brain images produced by LRE for the same age groups display salt-and-pepper noise. The detailed features of the brain images generated by the AMT for 74-year-olds and 84-year-olds appear blurred. Furthermore, the brain images generated by the WGAN are too bright. In contrast, the brain images of 74- and 84-year-olds produced by T-NIG are most similar to the corresponding original images. 

In terms of deformable characteristics, the brain images of 74-year-olds generated by each method closely resemble those of the original brain images. However, brain images of 84-year-olds generated by DRM, LRE, AMT, DIS, and WGAN exhibit undesirable deformable characteristics that differ from those of actual brain images. The deformable characteristics of the 84-year-old brain image predicted by T-NIG align well with those of the real brain image. T-NIG effectively captures the temporal texture features and temporal deformation features of brain images using TTCN and TDCN, thereby enhancing the model's ability to preserve disease-related features.

In summary, T-NIG exhibits superior performance in both short-term and long-term brain image prediction tasks.

\subsection{Long-term Prediction Among Different Age Groups}

To evaluate the performance of various methods for long-term brain image prediction in different age groups, we performed comparative experiments using the ADNI dataset. Our focus was on predicting MRI for individuals aged 50 to 89 years in the ADNI-1 cohort.

\subsubsection{Long-term brain image predict evaluation of different age groups} As shown in Table \ref{Table4}, T-NIG demonstrates the highest performance in long-term brain image prediction in all age groups. The performance of all methods declines with advancing age; notably, in the 80-89 age group, the performance of other methods is relatively low, while T-NIG significantly surpasses them. Compared to the second-best performing DIS model, T-NIG shows an improvement of 0.147 in SSIM and an increase of 3.3 in PSNR.

\begin{table}[htbp]
\centering
\caption{Long-term prediction of AD progress across different age groups(CN vs. MCI vs. AD) in ADNI-1.}
\setlength{\tabcolsep}{0.6mm}{} 
\begin{tabular}{c|ccc|ccc}
\hline
\multicolumn{1}{c|}{\multirow{2}{*}{Method}} & \multicolumn{3}{c|}{50-59 age group}              & \multicolumn{3}{c}{60-69 age group}               \\ \cline{2-7} 
\multicolumn{1}{c|}{}                        & SSIM$\uparrow$ & PSNR$\uparrow$ & MSE$\downarrow$ & SSIM$\uparrow$ & PSNR$\uparrow$ & MSE$\downarrow$ \\ \hline
DRM\cite{ref39}                                          & 0.764          & 30.8           & 0.771           & 0.706          & 26.4           & 0.898           \\
LRE\cite{ref40}                                          & 0.798          & 31.6           & 0.612           & 0.751          & 27.9           & 0.716           \\
AMT\cite{ref41}                                          & 0.835          & 32.9           & 0.496           & 0.810          & 28.8           & 0.571           \\
DIS\cite{ref7}                                          & 0.934          & 34.9           & 0.357           & 0.905          & 33.7           & 0.385           \\
WGAN\cite{ref43}                                         & 0.913          & 34.5           & 0.415           & 0.874          & 33.4           & 0.483           \\
T-NIG(Ours)                                  & $\mathbf{0.946}$          & $\mathbf{38.4}$           & $\mathbf{0.203}$           & $\mathbf{0.922}$          & $\mathbf{37.9}$           & $\mathbf{0.234}$           \\ \hline
\multicolumn{1}{c|}{\multirow{2}{*}{Method}} & \multicolumn{3}{c|}{70-79 age group}              & \multicolumn{3}{c}{80-89 age group}               \\ \cline{2-7} 
\multicolumn{1}{c|}{}                        & SSIM$\uparrow$ & PSNR$\uparrow$ & MSE$\downarrow$ & SSIM$\uparrow$ & PSNR$\uparrow$ & MSE$\downarrow$ \\ \hline
DRM\cite{ref39}                                          & 0.597          & 22.7           & 1.044           & 0.509          & 19.6           & 1.201           \\
LRE\cite{ref40}                                          & 0.683          & 23.4           & 0.856           & 0.582          & 20.3           & 0.998           \\
AMT\cite{ref41}                                          & 0.743          & 24.8           & 0.673           & 0.672          & 22.4           & 0.784           \\
DIS\cite{ref7}                                          & 0.830          & 31.4           & 0.427           & 0.751          & 29.9           & 0.487           \\
WGAN\cite{ref43}                                         & 0.818          & 30.2           & 0.551           & 0.733          & 28.7           & 0.643           \\
T-NIG(Ours)                                  & $\mathbf{0.912}$          & $\mathbf{35.1}$           & $\mathbf{0.275}$           & $\mathbf{0.898 }$         & $\mathbf{33.2}$           & $\mathbf{0.312}$           \\ \hline
\end{tabular}

\label{Table4}
\end{table}

\subsubsection{A detailed image generation comparison  of different age
groups: }
\begin{figure*}[htbp]
\centering
\centerline{\includegraphics[width=2\columnwidth]{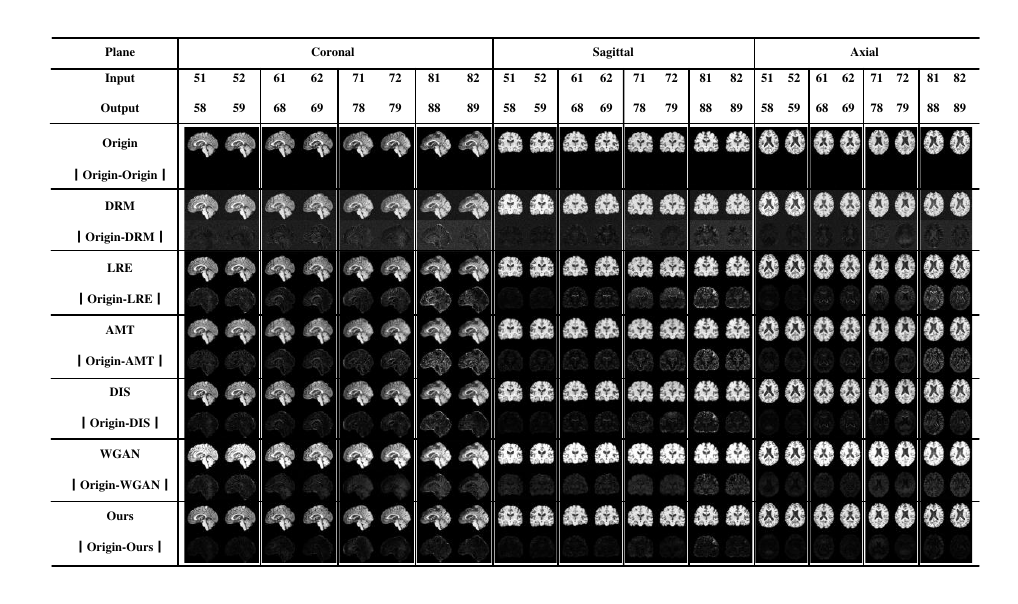}}
\caption{Brain image generation across different age groups. The first row represent the three planes in brain image; The second row illustrate the ages of the input and output brain images; The first row of each group displays the brain images generated by the corresponding models, while the second row of each group of images represents the differences between the generated brain images and the original images. }
\label{Figure_4}
\end{figure*}

Figure \ref{Figure_4} illustrates brain images generated by various methods in different age groups, ranging from 50s to 80s. 

In the 50-59 age group, the differences between brain images predicted by various methods and original images are relatively minor, as this group represents the early stages of disease progression. In particular,images produced by the DRM method display noise in both the brain region and the background. In contrast, the images generated by the T-NIG method exhibit the smallest deviation from the actual brain images. 

In the 60-69, 70-79 and 80-89 age groups, the discrepancies between brain images generated by various methods and the original images gradually increase with age. However, the difference between the brain images generated by T-NIG and the original images remains the smallest among all groups. In the 80 to 89 age group, the amount of background noise increases further in the images generated by the DRM method. Additionally, the LRE and AMT methods introduce more white noise throughout the entire brain, while DIS generates more noise at the edges of the brain, and WGAN produces a greater degree of blurriness. The experimental results indicate that T-NIG demonstrates stable performance in brain image prediction tasks across a wide range of age groups.

In summary, this experiment not only validates the effectiveness of T-NIG in long-term brain image prediction tasks, but also demonstrates its robustness in managing prediction tasks across various age groups.

\subsection{Performance of Models in Different Missing Ratio}

To investigate the impact of missing data on the models, Table \ref{Table6} presents three-class classifications using five ADNI-1 brain images with different missing ratios. When one brain image is absent, the missing rate is 20\%; when two brain images are absent, the missing rate is 30\%; when three brain images are absent, the missing rate is 60\%.

In Table \ref{Table6}, T-NIG achieves the best results in all missing ratios; the Acc decreases from 0.980 to 0.855 when using data with varying missing ratios, representing a 0.125 decrease. DIS, which ranks second, shows a decrease in Acc from 0.929 to 0.707, reflecting a decline of 0.202 under different missing ratios. The experimental results indicate that T-NIG effectively captures changes in the feature distribution among brain image sequences and enhances the quality of predicted brain images, even in the presence of inconsistent time intervals within these sequences.

\begin{table}[htbp]
\centering
\caption{Performance of Models in Different Missing Ratio.}
\setlength{\tabcolsep}{1mm}{} 
\begin{tabular}{c|cccccc}
\hline
\multirow{3}{*}{Method} & \multicolumn{6}{c}{Missing Ratio}                                   \\ \cline{2-7} 
                        & \multicolumn{3}{c|}{0\%}                     & \multicolumn{3}{c}{20\%} \\ \cline{2-7} 
                        & Acc$\uparrow$   & Pre$\uparrow$   & \multicolumn{1}{c|}{F1$\uparrow$}    & Acc$\uparrow$    & Pre$\uparrow$   & F1$\uparrow$    \\ \hline
DRM\cite{ref39}                     & 0.744 & 0.746 & \multicolumn{1}{c|}{0.749} & 0.694  & 0.701 & 0.687 \\
LRE\cite{ref40}                     & 0.785 & 0.784 & \multicolumn{1}{c|}{0.794} & 0.719  & 0.710 & 0.698 \\
AMT\cite{ref41}                     & 0.857 & 0.857 & \multicolumn{1}{c|}{0.855} & 0.756  & 0.741 & 0.749 \\
DIS\cite{ref7}                     & 0.929 & 0.928 & \multicolumn{1}{c|}{0.887} & 0.834  & 0.829 & 0.825 \\
WGAN\cite{ref43}                    & 0.911 & 0.907 & \multicolumn{1}{c|}{0.905} & 0.804  & 0.816 & 0.809 \\
T-NIG(Ours)             & $\mathbf{0.980}$ & $\mathbf{0.978}$ & \multicolumn{1}{c|}{$\mathbf{0.971}$} & $\mathbf{0.952}$  & $\mathbf{0.949}$ & $\mathbf{0.955}$ \\ \hline
\multirow{3}{*}{Method} & \multicolumn{6}{c}{Missing Ratio}                                   \\ \cline{2-7} 
                        & \multicolumn{3}{c|}{40\%}                    & \multicolumn{3}{c}{60\%} \\ \cline{2-7} 
                        & Acc$\uparrow$   & Pre$\uparrow$   & \multicolumn{1}{c|}{F1$\uparrow$}    & Acc$\uparrow$    & Pre$\uparrow$   & F1$\uparrow$    \\ \hline
DRM\cite{ref39}                     & 0.617 & 0.628 & \multicolumn{1}{c|}{0.607} & 0.537  & 0.552 & 0.543 \\
LRE\cite{ref40}                     & 0.607 & 0.613 & \multicolumn{1}{c|}{0.591} & 0.557  & 0.564 & 0.554 \\
AMT\cite{ref41}                     & 0.658 & 0.671 & \multicolumn{1}{c|}{0.663} & 0.616  & 0.624 & 0.629 \\
DIS\cite{ref7}                     & 0.758 & 0.746 & \multicolumn{1}{c|}{0.753} & 0.707  & 0.714 & 0.701 \\
WGAN\cite{ref43}                   & 0.756 & 0.742 & \multicolumn{1}{c|}{0.761} & 0.667  & 0.681 & 0.674 \\
T-NIG(Ours)             & $\mathbf{0.894}$ & $\mathbf{0.891}$ & \multicolumn{1}{c|}{$\mathbf{0.897}$} & $\mathbf{0.855}$  & $\mathbf{0.847}$ & $\mathbf{0.859}$ \\ \hline
\end{tabular}
\label{Table6}
\end{table}

\subsection{Ablation of Distribution}

 Distribution ablation experiments compare different distributions when modeling variations in brain imaging characteristics for the 10-year brain image prediction task in ADNI-1. The sequence of brain images corresponds to what is shown in Figure \ref{Figure_3}. The yellow distribution indicates changes in the features of the brain images.

\begin{figure*}[ht]
\centering
\centerline{\includegraphics[width=2\columnwidth]{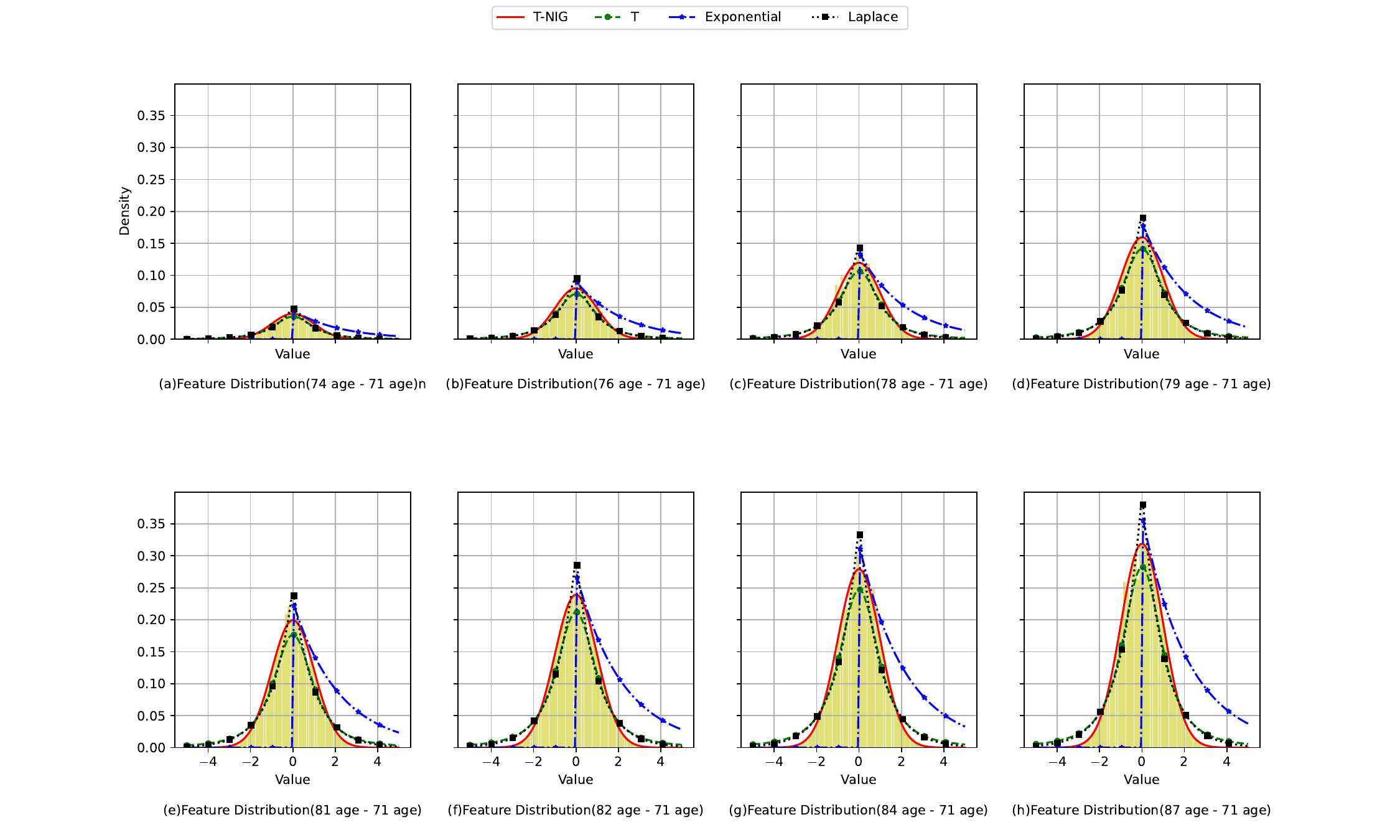}}
\caption{The experiment evaluates how well the T-NIG distribution, T distribution, Exponential distribution, and Laplace distribution fit the changes in features among brain images. The sequence of brain imaging follows the pattern illustrated in Figure \ref{Figure_3}.}
\label{Figure_5}
\end{figure*}

As illustrated in Figure \ref{Figure_5}, the T-NIG distribution aligns closely with the distribution of changes in brain imaging features. Although the fitting results for the T distribution are comparable to those of the T-NIG distribution, the T distribution is not as effective in accurately representing the mean and variance of the feature change distribution. In contrast, the Laplace distribution exhibits a notable variance difference compared to the actual distribution of changes in characteristics. Furthermore, when examining changes in thek feature distribution, both the mean and variance of the exponential distribution significantly deviate from the true distribution.

In general, although the fitting outcomes of the T distribution are quite similar to those of the T-NIG distribution, it does not adequately capture the mean and variance of the feature change distribution. The Laplace distribution shows a significant difference in variance, and the exponential distribution shows substantial discrepancies in both mean and variance compared to the actual distribution. 

\subsection{Ablation of Modules in T-NIG}

This section presents a stepwise ablation study of each module in the T-NIG framework for a 10-year brain image prediction task, with results detailed in Table \ref{Table7}. The experiments evaluate the temporal parameters based on the NIG distribution module (tNIG), the temporal texture feature extraction module (TTCN), the temporal deformation feature extraction module (TDCN), and the uncertainty estimation modules, including AL and EP, within the brain image prediction framework.

The findings indicate that each module of T-NIG positively influences the quality of brain image predictions. This study uses the ADNI-1 dataset to validate the effectiveness of the proposed T-NIG method in predicting brain image sequences over a 10-year period.

Compared to T-NIG, the ablation of the tNIG module results in a decrease in SSIM by 0.116, a reduction in PSNR by 5.1, and an increase in MSE by 0.433. Furthermore, ablation of the TTCN module leads to a decrease in PSNR by 3.6. The experimental results also indicate a decrease in PSNR of 3.8 compared to T-NIG. Furthermore, ablation of the AL and EP modules results in decreases in PSNR of 2.1 and 2.3, respectively, compared to T-NIG. When both the TTCN and TDCN modules are ablated simultaneously, the model experiences a decrease in PSNR of 0.135 compared to the complete T-NIG model. Lastly, simultaneous ablation of the AL and EP modules results in a PSNR decrease of 0.118 compared to the complete T-NIG model.

In summary, the ablation experiments conducted on each module demonstrate that the proposed tNIG, TTCN, TDCN, AL, and EP modules positively impact the enhancement of prediction quality in brain imaging tasks. Among these, the tNIG module has the most significant effect in improving the quality of the predicted brain images. Furthermore, the combined application of the TTCN and TDCN modules effectively enhances the prediction performance of the model. Similarly, the integration of the AL and EP modules significantly improves the model's predictive capabilities while reducing prediction errors.

\begin{table}[htbp]
\caption{Ablation of Modules in ADNI-1.}
\centering
\setlength{\tabcolsep}{1.5mm}{} 
\begin{tabular}{ccccc|ccc}
\hline
tNIG & TTCN                 & TDCN                 & AL                   & EP                    & SSIM $\uparrow$                     & PSNR$\uparrow$                     & MSE$\downarrow$   \\ \hline
    $\times$     &   $\sqrt{}$                     &   $\sqrt{}$                   &      $\sqrt{}$                &      $\sqrt{}$                 & $0.831$                     & 30.4                     & 0.764 \\
    
     $\sqrt{}$   &                  $\times$     &         $\sqrt{}$           &     $\sqrt{}$                 &           $\sqrt{}$            & 0.909                     & 31.9                     & 0.407 \\
     
   $\sqrt{}$    &          $\sqrt{}$          &         $\times$             &        $\sqrt{}$              &         $\sqrt{}$              & 0.897                     & 31.7                     & 0.419 \\
    
    $\sqrt{}$    & \multicolumn{1}{c}{$\sqrt{}$} & \multicolumn{1}{c}{$\sqrt{}$} & \multicolumn{1}{c}{$\times$} & \multicolumn{1}{c|}{$\sqrt{}$} & \multicolumn{1}{c}{0.910} & \multicolumn{1}{c}{33.4} & 0.394 \\
    
    $\sqrt{}$    & \multicolumn{1}{c}{$\sqrt{}$} & \multicolumn{1}{c}{$\sqrt{}$} & \multicolumn{1}{c}{$\sqrt{}$} & \multicolumn{1}{c|}{$\times$} & \multicolumn{1}{c}{0.907} & \multicolumn{1}{c}{33.2} & 0.401 \\
    
     $\sqrt{}$   & \multicolumn{1}{c}{$\sqrt{}$} & \multicolumn{1}{c}{$\sqrt{}$} & \multicolumn{1}{c}{$\times$} & \multicolumn{1}{c|}{$\times$} & \multicolumn{1}{c}{0.841} & \multicolumn{1}{c}{30.5} & 0.695   \\ 

 $\sqrt{}$   & \multicolumn{1}{c}{$\times$} & \multicolumn{1}{c}{$\times$} & \multicolumn{1}{c}{$\sqrt{}$} & \multicolumn{1}{c|}{$\sqrt{}$} & \multicolumn{1}{c}{0.824} & \multicolumn{1}{c}{26.3} &    0.752\\
     
     $\sqrt{}$   & \multicolumn{1}{c}{$\sqrt{}$} & \multicolumn{1}{c}{$\sqrt{}$} & \multicolumn{1}{c}{$\sqrt{}$} & \multicolumn{1}{c|}{$\sqrt{}$} & \multicolumn{1}{c}{$\mathbf{0.959}$} & \multicolumn{1}{c}{$\mathbf{37.2}$} & $\mathbf{0.302}$ \\ \hline
\end{tabular}
\label{Table7}
\end{table}

\section{DISCUSSION}
\label{sec:DISCUSSION}
In the healthcare sector, effectively predicting and tracking AD has been a considerable challenge for many years. Currently, predictions for Alzheimer's primarily depend on conventional brain imaging tests. However, irregularity in patient appointment schedules leads to a disjointed collection of brain images. The prediction technique T-NIG presented in this article can generate future brain images from these inconsistent time series. This development improves the ability to identify the evolving traits of AD, facilitating an earlier diagnosis of the disease.

The experiments demonstrated that T-NIG surpasses other techniques in preserving disease-related characteristics and is highly effective in predicting brain images over short and long durations in irregular time series. Ablation studies of each module reveal that utilizing both the TTCN and TDCN modules together enhances the model's prediction accuracy by identifying deformation features. Furthermore, by reducing errors, the integration of the AL and EP modules significantly enhances the quality of the predicted brain images, even when dealing with irregular time intervals.

In the analysis of the distribution within a sequence of brain images, the T-NIG distribution closely matches the observed variations in the sequential features of the brain imaging, surpassing other distributions in performance. Furthermore, experiments examining disease classifications of different image missing ratios reveal that T-NIG effectively captures changes in feature distribution throughout brain image sequences. An evaluation of disease-related characteristics through image generation experiments across various time frames and age groups indicates that the deformation patterns of brain images predicted by the T-NIG distribution closely resemble actual brain images when comparing images generated by different methods.

The experiments carried out to predict brain image sequences over periods of 5 and 10 years demonstrate that T-NIG surpasses other techniques in both short-term and long-term tasks. Furthermore, T-NIG exhibits exceptional performance in evaluations using the ADNI-1 and ADNI-2 datasets, showcasing its reliability across multiple datasets. In long-term prediction tests that include various age groups, T-NIG consistently provides reliable results in brain image prediction tasks across a wide age spectrum. 

Although the brain image prediction method has made some advances, there are still challenges that require further investigation. As the age of the subjects increases, the SSIM and PSNR values for predicting brain images tend to decrease, which corresponds to an increase in MSE, leading to a decrease in the model's performance when predicting extended sequences of brain images. Future studies will focus on enhancing the model's capability to predict these longer sequences.

\section{CONCLUSION}
\label{sec:conclusion}

T-NIG has created a predictive algorithm for MRI that utilizes a method to parameterize temporal image sequences based on the NIG distribution. This technique improves the algorithm's ability to maintain disease-related characteristics by incorporating both a temporal texture feature extraction module and a temporal deformation feature extraction module. By adding a temporal parameter to the NIG distribution, T-NIG enhances long-term brain image forecasts by understanding how distribution changes relate to time. Furthermore, through uncertainty estimation, T-NIG enhances the quality of brain image generation. Experimental findings show that T-NIG surpasses other algorithms in tasks related to brain image generation and disease state prediction.

\section{Acknowledgment }
\label{sec:Acknowledgment }
This work is supported in part by the Fujian Province Natural Science Foundation, China, under Grant 2022J01318; in part by the Huaqiao University High-Level Researchers Start-Up Fund Project under Grant 22BS105.

\section*{References}\bibliographystyle{ieeetr}
\bibliography{ref}

\end{document}